# New Fusion Algorithm provides an alternative approach to Robotic Path planning

**Ashutosh Kumar Tiwari**
BMS COLLEGE OF ENGINEERING, BANGALORE, INDIA
Email: reachatashutosh@gmail.com

**Sandeep Varma Nadimpalli**
ASSISTANT PROFESSOR, BMS COLLEGE OF ENGINEERING, BANGALORE, INDIA
Email: sandeepvarma.ise@bmsce.ac.in

*Abstract*— For rapid growth in technology and automation, human tasks are being taken over by robots as robots have proven to be better with both speed and precision. One of the major and widespread usage of these robots is in the industrial businesses, where they are employed to carry massive loads in and around work areas. As these working environments might not be completely localized and could be dynamically changing, new approaches must be evaluated to guarantee a crash-free way of performing duties.This paper presents a new and efficient fusion algorithm for solving path planning problem in a custom 2D environment. This fusion algorithm integrates an improved and optimized version of both, A* algorithm and the Artificial potential field method. Firstly, an initial or preliminary path is planned in the environmental model by adopting A* algorithm. The heuristic function of this A* algorithm is optimized and improved according to the environmental model. This is followed by selecting and saving the key nodes in the initial path Lastly, on the basis of these saved key nodes, path smoothing is done by artificial potential field method. Our simulation results carried out using Python viz. libraries indicate that the new fusion algorithm is feasible and superior in smoothness performance and can satisfy as a time-efficient and cheaper alternative to conventional A* strategies of path planning.

## I. INTRODUCTION

Path planning can be associated with the task of finding an optimized and crash-free path from one position called the source to another called the Goal, keeping a check on the obstacles, based on specific optimization rules. In simpler terms, it is a strategy of seeking the optimal path of movement for the robot from one point to another in space. It is also referred to as motion planning frequently as it caters the decision-making process of an object's motion in an environment. Robotic motion planning is of great significance in industrial as well as business scenarios and has drawn more and more attention from researchers. Therefore, research on motion planning is of great significance. An object that uses a path seeking algorithm to decide its traversing points in the space can be considered as autonomous in nature , and called a Robot. In such a context, path planning can be considered as a process of making discrete motions by the Robot for optimizing some entities by breaking down path movements into several iterative steps. Path planning problems are largely governed by the environment and can be classified as online and offline based on the nature of the workspace of the robot. In the former, the robot finds the position of the obstacle that continuously moves in the entire workspace by making use of real-time data acquiring types of equipment or sensors. In the later, pathfinding algorithms are used on the inputs, which is the data for the stationary obstacles with known geometry in the entire work area used by the robot. A lot of previous publications focus on these methods which are based on knowledge of environmental information. We need sophisticated modeling for the implementation of such concepts. In this paper, the conditional set up is that for a static 2D environment[1] therefore the fusion algorithm put in place is primarily offline.

Annealing algorithm[1], modified Artificial potential field, Generative Adversarial neural network method of path planning, particle swarm optimizations[2], grid method, framework space approach, A* algorithm, are some of the common and widely popular path planning algorithms. A* algorithm[3] is a minimum distance heuristic path searching algorithm which makes use of a heuristic function to calculate value at each node for optimized solutions. It provides the advantage of high processing speed and having much simpler operations for searching but suffers from certain limitations. It shows poor performance in smoothness and traceability of the traversed path as it is generally composed of straight-line segments. Moreover in cases where the distance between the start and the target position is more than the A* implementation can involve probing of too many nodes one after the other leading to the increase of running time.

Artificial potential field is a virtual force field-based method of local path planning[4]. It is identified by the construction of a combined potential field consisting of repulsive force field assumed to be set up by obstacles and attractive force field assumed to be set up by targets. The vector direction for the repulsive field is counter to the position of the obstacles and attractive field vector direction points towards the goal position. APF proves to be beneficial because of its simple structure, high gliding property and small calculations but suffers from certain limitations too. Reaching the equilibrium state at a certain point becomes a likely event resulting in a deadlock because of falling into the local optimal solution.

The paper takes the best of both the world of algorithms and integrates the benefits of both into one, called the fusion algorithm. Firstly, the optimized A* algorithm[5] reduces the overall path cost by calculating the path to goal directly without any feedback from the next frame. Secondly, the planned initial or preliminary path is divided into local goals or key nodes which are selected and saved to be addressed by the real-time reactive power of Artificial potential method, providing smoothness and better traceability with reduced time complexity.

## II. PROBLEM DESCRIPTION AND CONDITIONS

The path taken by the robot in the environment is simulated by establishing a static environment with obstacles of defined geometry and shape. Assumption of the fact that the robot could move in the work arena with ease is made for simulation.

### A. Inputs

The robot's 2D environment image is utilized and converted to a binary or bitmap image where obstacles are indicated by the black regions. The image is placed at the x & y-plane's origin and considered to be placed symmetrically about it. The coordinates of the source (starting point ) and the goal (target position ) are given as inputs.

### B. Optimized A* Algorithm

A* is a computer algorithm[6] estimating the minimal distance traversed by making use of the heuristic searching methodology[7]. The evaluation heuristic function which incorporates the information of heuristic used is expressed in (1):

$$(N) = g(N) + h(N) \quad (1)$$

where N is the present operating node on the path to be traversed, $g(N)$ is the source node to node N path cost, $h(N)$ is an evaluation heuristic function that represents cost estimates for the shortest path from node N to the goal. The aforementioned cost is the distance traversed by the robot.

The expanding of paths takes place step-by-step until one of its paths ends at the goal thereby constructing a tree of paths from the source node. To expand the adjacent nodes at each step the algorithm uses a heuristic function, using which it performs a repeated selection of nodes with minimum distance or cost estimates.

The heuristic used in the A* algorithm generally considers the Manhattan distance to calculate the distance between the current node to the goal node. The difference of the vertical and horizontal coordinates between the current node and goal node is calculated and the absolute sum is taken which is referred to as Manhattan distance. It is mathematically formulated as in (2) :

$$M_d = \left| x_{goal} - x_N \right| + \left| y_{goal} - y_N \right| \quad (2)$$

The A* algorithm to reach the goal node should never overestimate the actual cost and the evaluation function must satisfy the permissible condition. Manhattan distance does not satisfy the permissible condition under any situation. To satisfy the conditions of the environmental set up we have established a new heuristic also called evaluation function using the Euclidean distance. This helps to create an equilibrium between path accuracy and the processing speed. Such a heuristic function helps us to keep a check on both smaller and larger values of $h(N)$. The smaller $h(N)$ becomes the algorithm expands to more nodes at a slower speed but with great accuracy of the path. The larger the $h(N)$ becomes, a completely contrasting situation will occur. The actual distance from the current node to the goal node is referred to as Euclidean distance. It is mathematically formulated as in (3):

$$E_d = \sqrt{\left(x_{goal} - x_N\right)^2 + \left(y_{goal} - y_N\right)^2} \quad (3)$$

### C. Key Nodes/local goals in the planned path

A* algorithm is an offline algorithm that calculates the planned path from source to goal directly without any feedback from the next frames. The adjacent nodes in the preliminary or prior decided path are connected by straight lines. The paper adopts the selection of local nodes in the prior planned path and checks whether the line segments between the local nodes pass through the obstacles whose geometry and shape is defined in the input binary image. If not, the local nodes are saved and other unnecessary nodes are removed. As a result, an optimized planned path is obtained which can be made more smooth[8] and traceable[9,10] by making use of Artificial potential field.

### D. Optimized Artificial Potential field method

Artificial potential field[11,12,13] is used in global and local path planning. It is a classic approach for robot path planning which can be used for static or dynamic environments. It aims at finding a mathematical function for representing the energy of the system based on the ideas of physical potential field rules. It assumes that the repulsive and attractive forces exist where the attractive force is assumed to be set up between the robot and the goal and the one set up between the robot and the obstacles is the repulsive force. The general APF equation[14, 15] can be expressed as in (4) :

$$U(q) = U_{rep}(q) + U_{att}(q) \qquad (4)$$

Where $U_{rep}(q)$ is the repulsive function and $U_{att}(q)$ is the attractive function and the summation of both gives the total potential function used to control the robot. The prior planned path consists of a series of straight-line segments however it is not smooth at the turning point as the first order derivates are not steady and hence not continuous. This paper makes the line segments in the preliminary path as local goals to be achieved instead of directly applying the reactive power of APF from source to goal. For example, $S \rightarrow A \rightarrow B \rightarrow C \ldots \rightarrow G$, where $S$ is the source node and $G$ is the goal node. A local goal is established at node A by taking the next node B as the gravitational point with attractive function and all other obstacle points as repulsive points.

The gravitational function[16] is mathematically formulated as expressed in (5) :

$$F_{att} = G \vec{d}_g \qquad (5)$$

where G is gravitational constant, $\vec{d}_g$ is the cost between the current node to the local goal established.

The repulsive function[16] is formulated as in (6) :

$$F_{rep} = \begin{cases} F_{r_1} \vec{n}_0 + F_{r_2} \vec{n}_g & d_0 \leq \rho_0 \\ 0 & otherwise \end{cases} \qquad (6)$$

$$F_{r_1} = a \left(\frac{1}{d_0} - \frac{1}{\rho_0}\right) \frac{d_g^2}{d_0^2} \qquad (7)$$

$$F_{r_2} = \frac{na}{2} \left(\frac{1}{d_0} - \frac{1}{\rho_0}\right)^2 d_g^{n-1} \qquad (8)$$

where $a$ is the repulsive constant, $d_g$ is the cost between present operating node and the obstacles in the environment and $\rho_0$ is the overdone distance to an obstacle.

## III. METHODOLOGY

*A. Algorithmic flow*

The new Fusion algorithm presented in the paper can be expressed as in Figure 1. To verify the practicality and viability of the proposed algorithm, the simulation has been performed on binary image inputs (2D environment) using Python-OpenCV and other python graph enabling libraries. Since the paper deals with static 2D environments for robot path planning the obvious alternative of using MATLAB has been avoided [19].

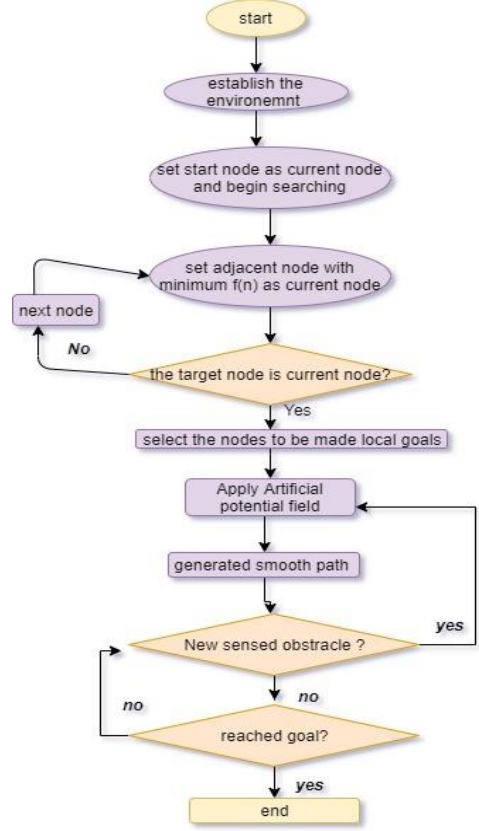

Fig.1. Algorithmic flow

*B. Proposed Fusion Algorithm*

The optimized A* algorithm[17,18] is used to find the preliminary path after which the APF[21, 22] is applied to reach the local goals on the A* planned path. This way we are using the best of both the worlds resulting in a smoother path and better traceability. We have named this algorithm as the Fusion algorithm expressed as below:

| Algorithm I. Fusion Algorithm: A* and APF |
|---|
| 1: local goals: = divide paths using A* Algorithm |
| 2: local source : = current source |
| 3: **for** $i \rightarrow 0$ to local goals |
| 4 : Path : = *empty list of entries* |
| 5 : **while** local source **is not** Final Goal **do** |
| 6 : Path : = Path +APF ( local source , local goal) |
| 7 : local source : = local goal |
| 8 : local goal : = next (local goals) |
| 9 : **end** , print(*running time & path length* ) |

III. RESULTS AND DISCUSSION

*A. Simulation Result and Analysis*

The proposed algorithm has been implemented in Python3 using python-OpenCV and other python graphing libraries. The various functions were employed in the python programming language. The input image consists of obstacles whose shape and geometry are defined. This is then transformed into a binary image. The program outputs the algorithm run time called the running time and the path length with the actual path traversed taken by the robot to move from source to goal. The code was initially run on Intel Core-i5-3470U @3.60 GHz with 16Gb RAM and NVIDIA IGPU where it produced remarkable results. The code was also run on Intel Pentium CPU 2117u @ 2.20 GHz showing the feasibility and the ability of the fusion algorithm to run successfully on slow processors with few resources, results of which are shown here in order to justify its applicability in such scenarios.

Table 1. Different cases and parameters used for simulation

| CASE | Coordinates of Initial position | Coordinates of final position |
|------|-------------------------------|------------------------------|
| 1 | source=[25,25] | goal=[180,280] |
| 2 | source=[25, 25] | goal=[280,340] |
| 3 | source =[25,25] | goal=[250,340] |
| 4 | source=[25,25] | goal=[330,250] |
| 5 | source=[25,25] | goal=[270,310] |
| 6 | source=[25,25] | goal=[320,170] |

Table 1. highlights the various simulation cases and parameters that were utilized for the demonstration of the working of the Fusion algorithm. The source is a pair of *(x, y)* coordinates which marks the initial or starting position for the pointer in our 2D environmental model. Pointer mimics the behaviour of a simple robot trying to move in a 2D space avoiding obstacles. We will, therefore, simply refer to it as a robot. The robot initiates its movement from the initial position, say [25,25] and move towards another *(x,y)* pair of coordinates known as the destination or goal. Here, the source and the goal coordinates are passed as parameters and provided explicitly for getting proper simulation results. Also, the position of the source is kept fixed for all the simulation cases for feasibility in comparison.

Table 2. Results for conventional A* Algorithm

| Simulation for Conventional A* Algorithm | |
|---|---|
| 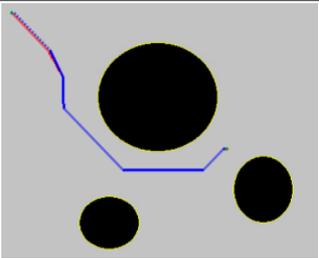 | **CASE 1** *Running time(s)*: 2.446611 *Path Length(cm)*: 353.4124 |
| 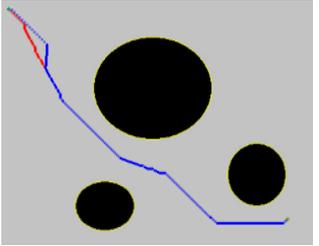 | **CASE 2** *Running time(s)*: 2.617699 *Path Length(cm)*: 487.1320 |
| 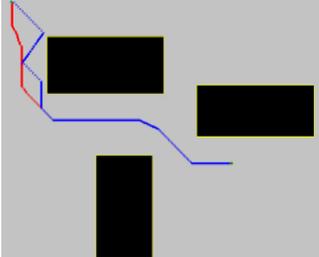 | **CASE 3** *Running time(s)*: 4.263914 *Path Length(cm)*: 459.2031 |
| 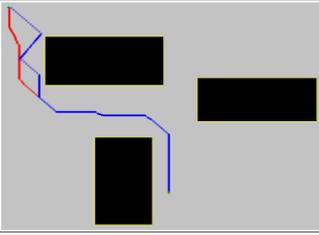 | **CASE 4** *Running time(s)*: 4.021216 *Path Length(cm)*: 458.0118 |
| 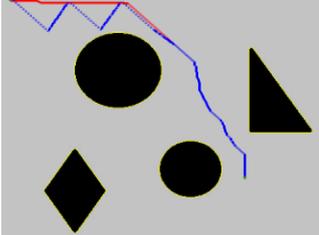 | **CASE 5** *Running time(s)*: 3.836856 *Path Length(cm)*: 512.3402 |
| 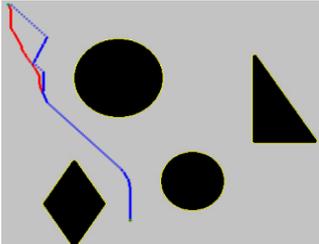 | **CASE 6** *Running time(s)*: 3.027221 *Path Length(cm)*: 372.7095 |

The 2D environment acts as a coordinate plane for the pointer with the plane being arranged in a quadrant where the *X*–axes is represented by the top and *Y*-axes is represented by the left edges respectively, of each image. The various geometrical shapes, shown in black, act as the obstacles which the robot is trying to avoid.

When the simulation begins, the robot plans its path at each timestep and tries to reach near the goal. At each step the robot tries to make a judgement of movement towards the goal and explores the plane. Red lines indicate the errors in judgement initially when the robot starts the exploration [20]. Noticeably, the blue and red line align considerably with slightly sharp corners due to the conventional A*algorithmic approach taken. Table 2. illustrates the simulation results for conventional A*algorithm with cases and parameters mentioned in Table 1.

Table 3. Results of proposed Fusion Algorithm

| Simulation of Fusion algorithm | |
|---|---|
| 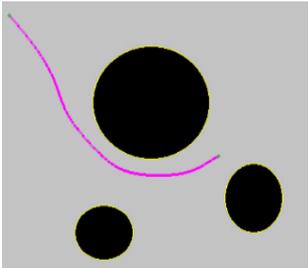 | **CASE 1** <br> Running time(s) : <br> 2.023923 <br> Path Length(cm): <br> 340.7240 |
| 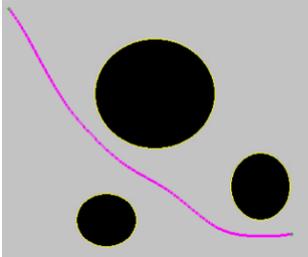 | **CASE 2** <br> Running time(s) : <br> 2.325824 <br> Path Length(cm) : <br> 471.9400 |
| 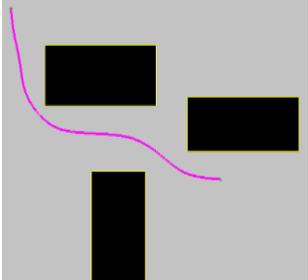 | **CASE 3** <br> Running time(s) : <br> 3.837976 <br> Path Length(cm) : <br> 446.6063 |
| 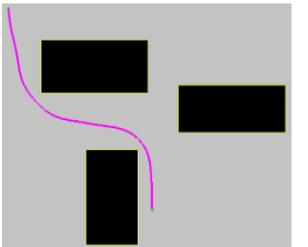 | **CASE 4** <br> Running time(s) : <br> 2.550975 <br> Path Length(cm) : <br> 432.9830 |
| 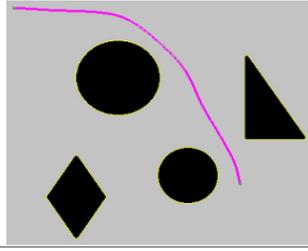 | **CASE 5** <br> Running time(s) : <br> 2.903379 <br> Path Length(cm) : <br> 497.2780 |
| 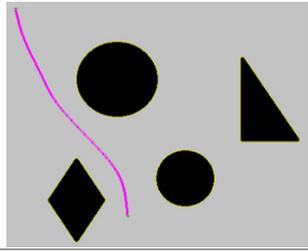 | **CASE 6** <br> Running time(s) : <br> 2.780669 <br> Path Length(cm) : <br> 372.8804 |

Table 3 is created from the simulation results, as in Tale 2, with the difference of use of proposed Fusion algorithm in the this case instead of the conventional A*algorithm.Noticeably, the comparison of static environment scenarios can be seen in Table 3 where the simulation results of the proposed algorithm for the same group of parameters has been illustrated. The path taken by the robot in the all the cases when using the proposed algorithm produces no sharp corners and is smooth from the source to the goal.

Both the tables, Table 2. And Table 3, show the paths along with the path length and running time when the robot moves from the source to the goal. The results of simulation and scenarios of path traversal lead to a comparison of running time, traceability and smoothness of the path undertaken by the robot. The case scenarios shown in Table 2. are substantially better than that of Table 2. in terms of better traceability and smoothness.

Table IV. Comparison statistics of running time and path length of A* Algorithm and Fusion Algorithm

| CASE | % reduction in running time(s) | % reduction in path length(cm) |
|---|---|---|
| 1 | 17.27 | 3.59 |
| 2 | 11.1 | 3.11 |
| 3 | 9.98 | 2.74 |
| 4 | 36.56 | 5.46 |
| 5 | 24.32 | 2.93 |
| 6 | 8.16 | 1.03 |

Table 4. further gives comparison statistics of results of the running time and path length for the fusion algorithm and the conventional A* Algorithm, shown

as a percentage change metric. Below are some of the observations,

- Along with better traceability and smoothness in the traversed path for all the above simulation cases, a considerable reduction in running time is observed along with a substantial decrease in the path length. This is a very remarkable result when working with slow processors.

- In few cases, both the running time and traversal path length is found to be significantly reduced thereby giving a path that is optimal and crash-free in the case of the proposed algorithm than the path traversed for conventional A* algorithm simulation cases.

- Sometimes, as in Case 6, for the same path length, the processing time calculated for the proposed algorithm is smaller than that of the time recorded for the conventional A* algorithm.This indicates that it is the optimal solution

## IV. CONCLUSION

This paper proposes a robot path planning algorithm based on the fusion of optimized A* algorithm and Artificial potential field method.The proposed algorithm is called Fusion algorithm. The viability of proposed algorithm is illustrated by simulations done on a static environment with fixed obstacles with specific geometries. Our simulations highlight considerable and competitive reduction in running time with minimal decrease of path length for the same source to goal coordinates. The results indicate that the Fusion algorithm is a time-efficient and cheaper alternative to the conventional A* algorithm of path planning. It also gives the added advantage of generating paths with better traceability and smoothness. The results therefore verify the superiority of the proposed algorithm over the conventional A* algorithm.

The present work creates abundant room for further progress which can be taken into consideration in the future investigations. The implementation of the proposed algorithm for local navigation of vehicles in a dynamic setting can be a prospect for future work. The proposed algorithm can be improvised with the use of GANs for unknown/partial information based path planning.

ACKNOWLEDGMENT

On the completion of this work, the authors would like to extend their gratitude to all whose help and advice led to the completion of this work. The author is very thankful to professor Camillo J. Taylor, Computer and Information science, University of Pennsylvania for his valuable help and advice during a MOOC course the author was enrolled.

## Authors' Profiles

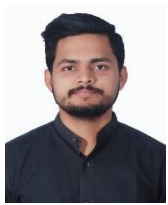

**Ashutosh Kumar Tiwari**, is currrently working as a Member Techincal Staff in Oracle Inc. He has completed his Bachelor of Engineering degree in Information Science and engineering from BMSCE, Banglore. He has interned at R&D labs of two major MNC's, Accenture and Epicor Incorporation and worked on Deep learning and reinforcement learning algorithms to drive business goals. He is an active member of Google developer group Bengaluru. His research interests include ML, DL , Data Engineering and Cloud Computing .

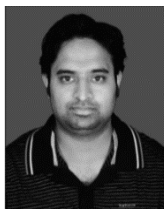

**Sandeep Varma Nadimpalli,** is currently working as Assistant Professor since 2014 in the Department of Information Science and Engineering, B.M.S College of Engineering. He received his B.Tech. degree in Information Technology from JNTU Hyderabad, Telangana, India in 2007. He received his M.Tech. from Andhra University in 2009 and his Ph.D. in computer science and systems engineeringfrom Andhra University in 2015. He also worked as Junior Research Fellow (Professional) from 2010 to 2011 and later worked as Senior Research Fellow from 2011 to 2013 at Andhra University. His research interests include Data engineering, Data Privacy, Cloud Computing and Social Networks. He is a member of IEEE.